\documentclass{article}
\usepackage{ijcai25}

\usepackage{times}
\usepackage{soul}
\usepackage{url}
\usepackage[hidelinks]{hyperref}
\usepackage[utf8]{inputenc}
\usepackage[small]{caption}
\usepackage{graphicx}
\usepackage{amsmath}
\usepackage{amsthm}
\usepackage{amssymb}
\usepackage{booktabs}
\usepackage{algorithm}
\usepackage{algorithmic}
\usepackage[switch]{lineno}

\title{Multi-Source Collaborative Style Augmentation and Domain-Invariant Learning for Federated Domain Generalization}

\author{
    Yikang Wei
    \affiliations
    College of Intelligence and Computing, Tianjin University, Tianjin, China
    \emails
    yikang@tju.edu.cn
}

\begin{document}

\maketitle

\begin{abstract}
    Federated domain generalization aims to learn a generalizable model from multiple decentralized source domains for deploying on the unseen target domain. The style augmentation methods have achieved great progress on domain generalization. However, the existing style augmentation methods either explore the data styles within isolated source domain or interpolate the style information across existing source domains under the data decentralization scenario, which leads to limited style space. To address this issue, we propose a Multi-source Collaborative Style Augmentation and Domain-invariant learning method (MCSAD) for federated domain generalization. Specifically, we propose a multi-source collaborative style augmentation module to generate data in the broader style space. Furthermore, we conduct domain-invariant learning between the original data and augmented data by cross-domain feature alignment within the same class and classes relation ensemble distillation between different classes to learn a domain-invariant model. By alternatively conducting collaborative style augmentation and domain-invariant learning, the model can generalize well on unseen target domain. Extensive experiments on multiple domain generalization datasets indicate that our method significantly outperforms the state-of-the-art federated domain generalization methods.
\end{abstract}

\section{Introduction}

Domain generalization is a challenging task in machine learning, which involves training a model on source domains to generalize well on the unseen target domain. As the data from different domains exist domain shift\cite{huang2024federated,wu2024prototype,li2024prompt,tang2024bootstrap,tang2025ocrt}, the model trained on source domains tends to be domain-specific, leading to the degraded performance on unseen target domain. The domain-specific bias between source domains and the unseen target domain partially stems from the differences of data styles. Based on this assumption, the existing domain generalization methods diversify the style of source domains to against the domain-specific bias, e.g. some Single-Domain Generalization (Single-DG) methods\cite{kang2022style,zhou2020learning,zhou2020deep} expand the style space of single source domain and the Multi-Domain Generalization (Multi-DG) methods\cite{xu2021fourier,zhou2021mixstyle} conduct style interpolation between multiple source domains to generate data with novel styles.

\begin{figure}[htb]
    \centering
    \includegraphics[width=0.48\textwidth]{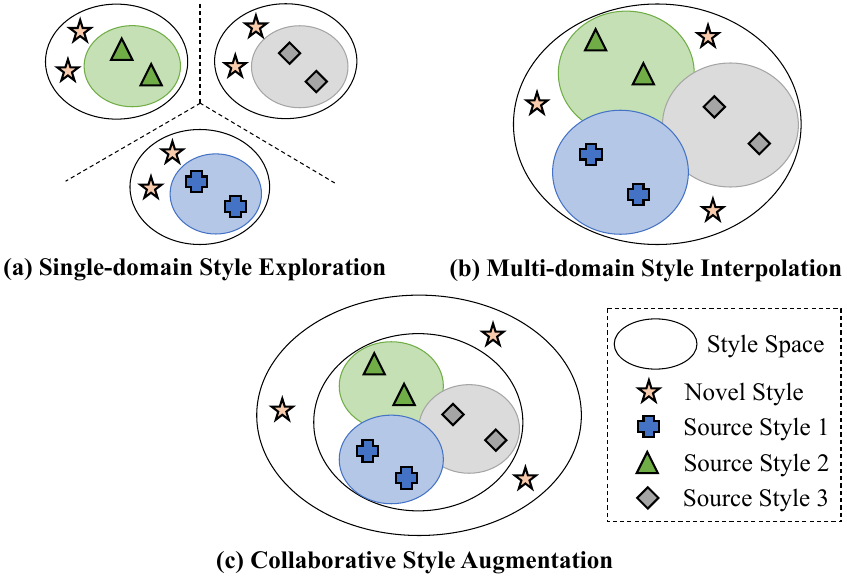}
    \caption{(a) Single-domain style exploration on the isolated source domains ignores the styles of other domains, which leads to limited style diversity. (b) Multi-domain style interpolation mixes the shared style information across decentralized source domains, which leads to the generated styles within the existing source domains. (c) Collaborative style augmentation proposed in this work generates data with novel styles out of the existing source domains, which can explore the broader style space.}
    \label{fig2}
\end{figure}

Conventional Multi-DG\cite{dou2019domain,du2022cross,zhou2021mixstyle,Tang_2024_BMVC} assumes that the data from different source domains are centralized and available for learning a domain-invariant model. However, considering the data privacy and communication cost, the data from different source domains are decentralized and forbidden to share across domains\cite{yuan2023collaborative,wu2021collaborative,9940295,wei2023exploring,wei2022dual}. In this work, we consider the Federated Domain Generalization (FedDG) scenario\cite{yuan2023collaborative,wei2024multi,park2023stablefdg,xu2023federated}, where the data from different source domains are decentralized.

Under the data decentralization scenario, either single-domain style exploration within each isolated source domain or multi-domain style interpolation across decentralized source domains, could be used to diversify the source domains. However, the diversity of styles generated by existing methods are still limited. Firstly, the single-domain style exploration methods \cite{zhong2022adversarial,wang2021learning} employed by Single-DG ignore the styles of other source domains, which degrades the generalization performance, as shown in Figure \ref{fig2} (a). Secondly, the multi-domain style interpolation methods \cite{liu2021feddg,chen2023federated} used by Multi-DG result in limited styles within the existing source domains, as shown in Figure \ref{fig2} (b). Furthermore, the style information across decentralized source domains should be shared for conducting the multi-domain style interpolation under the data decentralization scenario, which have the risk of privacy leakage and lead to additional communication cost. How to explore the out-of-distribution styles between multiple decentralized source domains becoming the key challenge.

For exploring the broader style space across multiple decentralized source domains, we propose a FedDG method to collaboratively explore the out-of-distribution styles across decentralized source domains. Inspired by the existing methods \cite{huang2017arbitrary,zhong2022adversarial}, we conduct channel-wise statistic features transformation to explore the out-of-distribution styles with the collaboration of other source domains. Specifically, the classifier heads from other source domains are used as the discriminators. The augmented data which cannot be correctly classified by the existing classifiers from other domains, tend to be out-of-distribution. Different from the existing single-domain exploration methods and multi-domain style interpolation methods, the proposed Collaborative Style Augmentation (CSA) method can collaboratively explore the broader style space between the decentralized source domains, as shown in Figure \ref{fig2}(c). And the risk of privacy leakage and the cost of communication are smaller than sharing style information across decentralized domains \cite{liu2021feddg,chen2023federated}.

Furthermore, we conduct domain-invariant learning between the original data and augmented data to learn the domain-invariant information for generalizing well on the unseen target domain. Specifically, we align the features of original data and augmented data by contrastive loss to increase the compactness of representations within the same class. And we distill the classes relationships from multiple classifier heads to improve the generalizable ability of model. On the decentralized source domains, the collaborative style augmentation and domain-invariant learning are conducted alternatively to improve the generalizable ability of models on the unseen target domain.

The experiments on multiple domain generalization datasets indicate the effectiveness of our method. The major contributions of this work can be summarized as follows:

\begin{itemize}
  \item We propose a Collaborative Style Augmentation module to explore the out-of-distribution styles under the data decentralization scenario with the collaboration of other source domains.
  \item We conduct domain-invariant learning between the original data and augmented data by contrastive alignment and ensemble distillation for learning domain-invariant model which can generalize well on the unseen target domain.
  \item The results and analysis on multiple domain generalization datasets indicate that our method outperforms the state-of-the-art FedDG methods significantly.
\end{itemize}

\section{Related Works}

\textbf{Multi-source Domain Generalization.} Multi-DG aims to learn a generalizable model by utilizing multiple labeled source domains. Conventional Multi-DG methods can be categorized into (1) data augmentation methods, (2) domain-invariant representation learning methods, and (3) other learning strategies. Data augmentation methods aim to diversify the source domains for improving the generalization ability of model on unseen target domain. For example, L2A-OT \cite{zhou2020learning} and DDAIG \cite{zhou2020deep} learn the image generator to generate images with novel styles, FACT \cite{xu2021fourier}, MixStyle \cite{zhou2021mixstyle}, and EFDMix \cite{zhang2022exact} conduct style interpolation between different source domains to generate data with novel styles. Domain-invariant representation learning methods aim to learn the intrinsic semantic representation from multiple source domains for applying to unseen target domain. Such as the self-supervised learning methods, JiGen \cite{carlucci2019domain} and EISNet \cite{wang2020learning} utilize the Jigsaw auxiliary task to learn the domain-invariant representations. And RSC \cite{huang2020self}, CDG \cite{du2022cross}, I$^{2}$-ADR \cite{meng2022attention}, and DomainDrop \cite{guo2023domaindrop} learn the intrinsic semantic representations by removing the spurious correlation features. Learning strategies, e.g. MASF \cite{dou2019domain} utilizes meta-learning to learn intrinsic semantic representations across different domains. And DAEL \cite{zhou2021domain} utilizes ensemble learning to learn the complementary knowledge from different domains. However, these conventional Multi-DG methods assume that the data from multiple source domains can be accessed simultaneously, which cannot be satisfied under the data decentralization scenario.

\begin{figure*}[htb]
    \centering
    \includegraphics[width=1.0\textwidth]{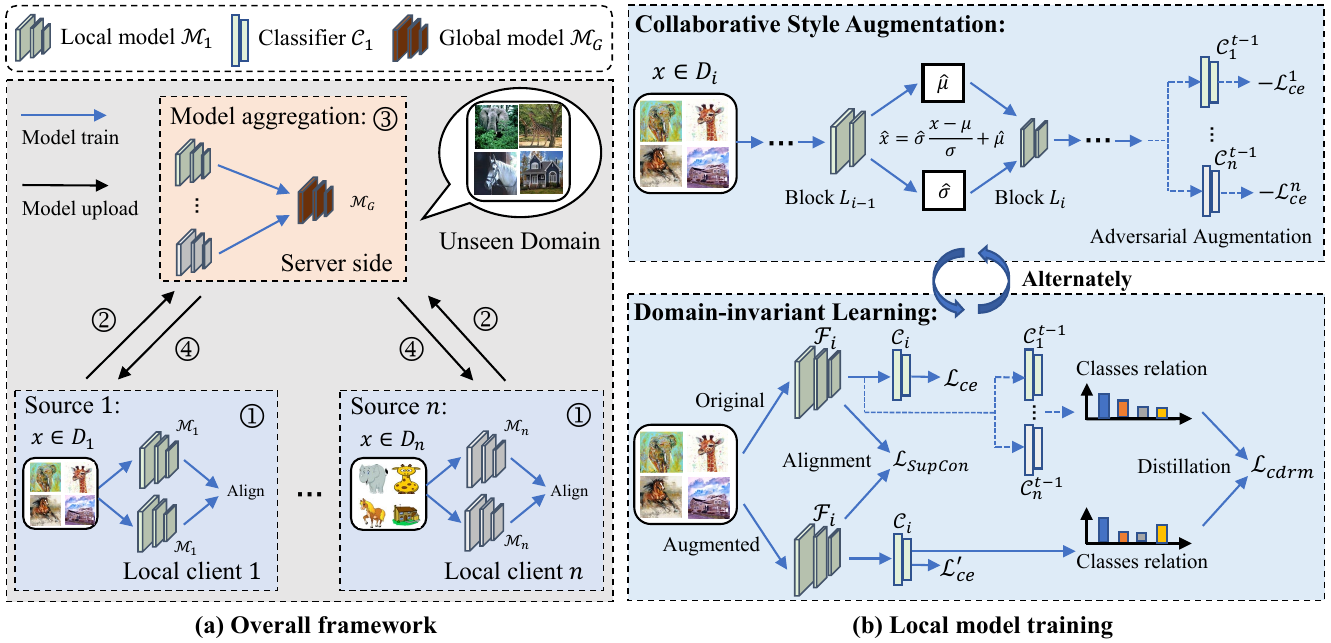}
    \caption{(a) The overall framework of our method. The local source domain models $\{\mathcal{M}_{i}\}_{i=1}^{n}$ are trained on the isolated local clients and aggregated on the server side to obtain the global model $\mathcal{M}_{G}$, which will be used on the unseen domain for better generalizable ability. (b) The local model training on decentralized source domains, where collaborative style augmentation and domain-invariant learning are conducted alternatively to improve the generalizable ability of local source domain models.}
    \label{fig3}
\end{figure*}

\textbf{Single-source Domain Generalization.} Single-DG directly learns a generalizable model from single source domain. The existing Single-DG methods \cite{wang2021learning,li2021progressive,xu2023simde,zhou2021mixstyle,wang2021learning} usually synthesize images or features with novel styles and keep the semantic information invariant, which can expand the diversity of source domain to achieve better generalization ability. For example, StyleNeophile \cite{kang2022style} and DSU \cite{liu2021feddg} explore novel styles in feature space to expand the source domain. And the AdvStyle \cite{zhong2022adversarial} generates images with novel styles by adversarial augmentation. However, these Single-DG methods cannot collaboratively utilize multiple decentralized source domains and lead to limited generalization performance. We will give a detailed comparison and analysis in experiments by applying these single-domain style exploration methods on FedDG scenario.

\textbf{Federated Domain Generalization.} FedDG collaboratively train multiple decentralized source domains for obtaining a model generalizing well on the unseen target domain. The existing FedDG methods utilize the federated learning framework e.g. FedAvg \cite{pmlr-v54-mcmahan17a} to collaboratively train the local source models on the decentralized source domains and aggregate the local source models on the server side. The existing FedDG methods can be categorized into (1) domain-invariant learning methods and (2) model aggregation strategies. The domain-invariant learning methods focus on exploring the data with out-of-distribution styles for generalizing on the unseen target domain. For example, ELCFS \cite{liu2021feddg} and CCST \cite{chen2023federated} interpolate the shared style information across decentralized source domains to generate data with novel styles, the StableFDG \cite{park2023stablefdg} explores novel styles based on the shared style information across decentralized source domains. And FADH \cite{xu2023federated} trains additional image generators on the local clients to generate images with novel styles. COPA \cite{wu2021collaborative} augments the images by pre-defined augmentation pool, e.g. RandAug \cite{cubuk2020randaugment} for learning the domain-invariant model. Different from these multi-domain style interpolation methods, our method can explore the broader style space without sharing the style information across decentralized source domains. The model aggregation strategies e.g. CASC \cite{yuan2023collaborative} and GA \cite{zhang2023federated} calibrate the aggregation weights of local models on server side to obtain a fairness global model for the better generalization ability. Different from these methods, our method aims to learn domain-invariant model on local clients.

\section{Method}

\textbf{Problem Definition.} We focus on the generic image classification task for FedDG. Given multiple decentralized source domains $\{X_{i},Y_{i}\}_{i=1}^{n}$, each domain $\{X_{i},Y_{i}\}$ contains $N_{i}$ samples. Data from different source domains are located on the isolated clients and forbidden to share across clients. The goal of FedDG is to learn a generalizable global model $\mathcal{M}_{G}$ from multiple decentralized source domains $\{X_{i},Y_{i}\}_{i=1}^{n}$ for deploying on the unseen target domain ${X_{t}}$. The different source domains $\{X_{i},Y_{i}\}_{i=1}^{n}$ and unseen target domain ${X_{t}}$ exist the covariate shift, e.g. the marginal distribution of images $P(X)$ differs but the conditional label distribution $P(Y|X)$ keeps same across domains.

\subsection{Overall Framework}
We propose a multi-source collaborative style augmentation and domain-invariant learning method to explore out-of-distribution styles and learn the domain-invariant model across decentralized source domains. As shown in Figure \ref{fig3}(a), each client contains a source domain, there are four steps to collaboratively train the decentralized source domains. On Step 1, the local models $\{\mathcal{M}_{i}\}_{i=1}^{n}$ are trained on the isolated source domains respectively. Each local model $\mathcal{M}_{i}$ contains a feature extractor $\mathcal{F}_{i}$ and a classifier head $\mathcal{C}_{i}$. On Step 2, the local models $\{\mathcal{M}_{i}\}_{i=1}^{n}$ are uploaded to the server side. On Step 3, the local models $\{\mathcal{M}_{i}\}_{i=1}^{n}$ are aggregated by parameter averaging to obtain a global model $\mathcal{M}_{G}$. Then, the global model $\mathcal{M}_{G}$ and the classifier heads $\{\mathcal{C}_{i}^{t-1}\}_{i=1}^{n}$ of different source domains are broadcasted to local clients on Step 4, which are used to conduct local training with the collaboration of other domains on the next t-th round training. The four steps are conducted iteratively until the global model convergence. Finally, the global model $\mathcal{M}_{G}$ is deployed on the unseen domain $\{X_{t}\}$.

For learning a domain-invariant model from multiple decentralized source domains to generalize well on unseen target domain, we propose (1) Collaborative Style Augmentation (CSA) to explore the out-of-distribution styles with the collaboration of other domain classifier heads, and (2) domain-invariant learning between the original data and augmented data to learn the intrinsic semantic information within class by contrastive alignment and the relationship between classes by cross-domain relation matching.

\subsection{Collaborative Style Augmentation}
Inspired by the existing style augmentation methods, e.g. MixStyle\cite{zhou2021mixstyle}, the style information of the feature $f\in\mathbb{R}^{H\times W \times C}$ with the spatial size $H\times W$ can be revealed by the channel-wise mean $\mu$ and standard deviation $\sigma$:

\begin{equation}\label{eq1}
  \mu = \frac{1}{HW}\sum_{h\in H,w\in W}{f_{h,w}},
\end{equation}

\begin{equation}\label{eq2}
  \sigma = \sqrt{\frac{1}{HW}\sum_{h\in H,w\in W}{(f_{h,w}-\mu)^{2}}}.
\end{equation}

For generating features with novel style, the original feature $f$ is normalized by the channel-wise mean $\mu$ and standard deviation $\sigma$, then the normalized feature is scaled by the novel standard deviation $\hat{\sigma}$ and added by the novel mean $\hat{\mu}$:

\begin{equation}\label{eq3}
  \hat{f} = \hat{\sigma}\frac{f-\mu}{\sigma} + \hat{\mu}.
\end{equation}

For generating novel statistic mean $\hat{\mu}$ and standard deviation $\hat{\sigma}$, the existing style augmentation methods e.g. single-domain style exploration method DSU \cite{li2021uncertainty} expands the mean $\mu$ and standard deviation $\sigma$ by sampling perturbs from the estimated Gaussian distribution, the AdvStyle \cite{zhong2022adversarial} expends $\mu$ and $\sigma$ by maximizing the cross-entropy loss of the current model $\mathcal{F}_{i}\circ \mathcal{C}_{i}$ to learn the adversarial perturbs. However, these single-domain style exploration methods only explores the styles based on the current source domain and ignores the other decentralized source domains, which leads to limited diversity of styles. Although some FedDG methods e.g. CCST \cite{chen2023federated} and StableFDG \cite{park2023stablefdg} share the mean $\mu$ and standard deviation $\sigma$ across multiple decentralized source domains to conduct multi-domain style interpolation, the generated novel styles still come from the style space of existing source domains.

Different from these single-domain exploration methods and multi-domain interpolation methods, we propose collaborative style augmentation to generate the out-of-distribution styles with the collaboration of other source domains. Specifically, we use the classifier heads from other source domains as the discriminators to guide the generalization of novel styles, as shown in the top of Figure \ref{fig3}(b). On the isolated domain $D_{i}$, the style statistics $\hat{\mu}$ and $\hat{\sigma}$ can be learned as follows:

\begin{equation}\label{eq6}
  \hat{\mu} = \mu + \frac{1}{n}\sum_{j=1}^{n}\eta\nabla_{\mu}\mathcal{L}_{ce}^{'}{(\mathcal{F}_{i}^{l}\circ\mathcal{C}_{j}^{t-1};\hat{f})},
\end{equation}

\begin{equation}\label{eq7}
  \hat{\sigma} = \sigma + \frac{1}{n}\sum_{j=1}^{n}\eta\nabla_{\sigma}\mathcal{L}_{ce}^{'}{(\mathcal{F}_{i}^{l}\circ\mathcal{C}_{j}^{t-1};\hat{f})}.
\end{equation}

The $\mathcal{L}_{ce}^{'}=-y\log(\delta(\mathcal{F}_{i}^{l}\circ\mathcal{C}_{j}^{t-1}(\hat{f})))$, $\delta$ is the softmax function. $\{\mathcal{C}_{j}^{t-1}\}_{j=1}^{n}$ are the classifier heads of decentralized source domains from the last round $t-1$. For the feature $f$ extracted by hiden layers, we conduct adversarial style augmentation by using the rest neural network layers $\mathcal{F}_{i}^{l}$. In experiments, we analysis where are the best location to conduct feature augmentation.

In Equation \ref{eq6} and Equation \ref{eq7}, the generated novel styles tend to be away from the decision boundary of existing classifiers from different source domains, so that the features with novel styles are out of the existing source domains.

%

\subsection{Domain-invariant Learning}
The original data and augmented data are used to train the local model with cross-entropy:

\begin{equation}\label{task_loss}
  \mathcal{L}_{task} = \frac{1}{2}(\mathcal{L}_{ce}(\mathcal{F}_{i}\circ\mathcal{C}_{i};x) + \mathcal{L}_{ce}^{'}(\mathcal{F}_{i}^{l}\circ\mathcal{C}_{i};\hat{f})).
\end{equation}

For further improving the generalization ability of model, we conduct domain-invariant learning between original data and augmented data. Firstly, the cross-domain feature alignment is utilized to learn the compact representations within the class by contrastive alignment. Secondly, Cross-Domain Relationship Matching (CDRM) is proposed to learn the relationship between classes from the ensemble of multiple classifier heads.

\subsubsection{Cross-domain feature alignment}

We conduct supervised contrastive alignment \cite{khosla2020supervised} at feature level to improve the compactness of representations within the same class. For the $i$-th isolated domain:

\begin{equation}\label{eq10}
  \mathcal{L}_{SupCon} = -\sum_{j=0}^{2N_{i}}{\frac{1}{|P(j)|}}\sum_{p\in P(j)}{\log \frac{e^{(z_{j}\cdot z_{p}/\tau)}}{\sum_{a\in A(j)}{e^{(z_{j}\cdot z_{a}/\tau)}}}}.
\end{equation}

The $z_{j}$ are the feature of $j$-th image extracted by feature extractor $\mathcal{F}_{i}$. The $P(j)$ is the set of features with the same label as $j$-th image, $|P(j)|$ indicates the number of features in set $P(j)$. $A(j)$ indicates the features of original data and augmented data. $\tau$ is the temperature parameter, hear we set 0.07 following previous work.

\subsubsection{Cross domain relation matching}

Furthermore, we conduct cross-domain relation matching to keep the intrinsic similarity between classes, e.g. the category of giraffe has a larger similarity with the elephant than the house. For obtaining the intrinsic relationship between classes, we calculate the ensemble logits $l_{ens}$ of the same class from multiple classifiers $\{\mathcal{C}_{i}^{t-1}\}_{i=1}^{n}$. For example, on the $m$-th source domain, we calculate the $l_{ens}^{k}$ of class $k$ by averaging the logits of original images from different classifier heads within the same class $C(k)$:

\begin{equation}\label{eq13}
    l_{ens}^{k} = \sum_{i=1}^{n}\sum_{j\in C(k)}{\mathcal{F}_{m}\circ\mathcal{C}_{i}^{t-1}(x_{j})},
\end{equation}

then $l_{ens}^{k}$ is scaled with a temperature $\tau=2.0$ by softmax function for smoothing the probability between classes, which can capture intrinsic semantic relationship between current class $k$ and other classes. The smoothed $l_{ens}^{k}$ of different classes are $\{p_{ens}^{k}\}_{k=1}^{K}$.

We also calculate the class relation $\{p_{cur}^{k}\}_{k=1}^{K}$ of the original images and $\{\hat{p}_{cur}^{k}\}_{k=1}^{K}$ of the stylized images from current model $\mathcal{F}_{m}\circ\mathcal{C}_{m}$, which are used to match the ensemble class relation $\{p_{ens}^{k}\}_{k=1}^{K}$:

\begin{equation}\label{eq11}
  \mathcal{L}_{cdrm} = \sum_{k=1}^{K}\sum_{j=1}^{K}({p_{ens}^{k(j)}\log p_{cur}^{k(j)}} + {p}_{ens}^{k(j)}\log \hat{p}_{cur}^{k(j)}),
\end{equation}

The overall loss of semantic learning on local clients is as follows:

\begin{equation}\label{local_loss}
  \mathcal{L}_{local} = \mathcal{L}_{task} + \lambda_{con}\mathcal{L}_{con} + \lambda_{cdrm}\mathcal{L}_{cdrm},
\end{equation}

where $\lambda_{con}$ and $\lambda_{cdrm}$ are the hyper-parameters to balance different losses.

\subsection{Model Aggregation}

On the server side, we aggregate different source domain models $\{\mathcal{M}_{i}\}_{i=1}^{n}$ by parameter averaging:

\begin{equation}\label{model_agg}
  \mathcal{M}_{G} = \sum_{i=1}^{n}{\frac{N_{i}}{N_{totoal}}\mathcal{M}_{i}},
\end{equation}

the $N_{total}$ is the sum of all samples from different source domains. The aggregated global model $\mathcal{M}_{G}$ is used as the initial local model on local clients for the next round of training.

\section{Experiments}

\subsection{Datasets}
In this work, we follow the previous DG works \cite{zhou2021mixstyle,wu2021collaborative} to evaluate our method on three image classification datasets: \textbf{PACS}, \textbf{Office-Home}, and \textbf{VLCS}. \textbf{PACS} contains 9,991 images of 7 categories from four domains, Art-Painting (Art), Cartoon, Photo, and Sketch. Following previous works \cite{zhou2021mixstyle}, we split the data of each domain into 80\% for training and 20\% for testing. \textbf{Office-Home} contains 15,500 images of 65 categories from four domains, Artistic (Art), Clipart, Product, and Real-World (Real). Following the previous works \cite{zhou2021mixstyle}, we split each domain into 90\% as the training set and 10\% as the test set. \textbf{VLCS} contains 10,729 images of 5 categories from four domains, Pascal, LabelMe, Caltech, and Sun. Following previous works \cite{chen2023federated}, we split the data of each domain into 80\% for training and 20\% for testing.

We utilize the leave-one-domain-out protocol \cite{zhou2021mixstyle} to select a domain as the unseen target domain for evaluation and the rest as the source domains for training.

\subsection{Implementation Details}

Following previous works \cite{yuan2023collaborative,wu2021collaborative}, we use the pre-trained ResNet-18 on ImageNet as the backbone for PACS, Office-Home, and VLCS dataset. The SGD optimizer is used to optimize the network with momentum 0.9 and weight decay 5e-4. The initial learning rate is 0.001 decayed by the cosine schedule to 0.0001 for the PACS and VLCS datasets. For the Office-Home dataset, the initial learning rate is 0.002 and decayed to 0.0001. The batch size is 16 for PACS and VLCS, and 30 for Office-Home. The adversarial learning rate $\eta$ is 1.0 for PACS and VLCS dataset, 0.3 for Office-Home datasets. The $\lambda_{SupCon}$ is 1.0 for PACS and Office-Home dataset, 0.3 for VLCS dataset. The $\lambda_{cdrm}$ is 4.0 for PACS, 0.7 for Office-Home, and 0.3 for VLCS dataset. the $\tau$ of $\mathcal{L}_{cdrm}$ is 1.5 for all dataset. The values of hyper-parameters are set according to the performance on validation set of source domains.

We train the local model on each client 1 epoch in Step 1 and then upload the local models to the server side for model aggregation. The total communication rounds between clients and server is 40 for PACS, Office-Home, VLCS. All experiments are conducted three times with different random seeds, and the mean accuracy (\%) is reported.

For evaluating the effectiveness of our method, we make a comparison with the state-of-the-art DG methods. The DeepAll indicates training the model by cross-entropy on the centralized source domains. The FedAvg \cite{pmlr-v54-mcmahan17a} indicates collaborative training the decentralized source domains. As shown in Table \ref{pacs}, Table \ref{office_home}, and Table \ref{vlcs}, we report the accuracy (\%) of state-of-the-art methods under the data centralized and decentralized scenario, where the \textbf{Dec.} indicates the data decentralization scenario.

\begin{table}[htbp]
\centering
  \scalebox{0.78}{
  \setlength{\tabcolsep}{1.0mm}{
  \begin{tabular}{lcccccc}
  \hline
  Methods & Dec. & Art & Cartoon & Photo & Sketch & Avg \\
  \hline
  DeepAll \cite{zhou2021domain}   & $\times$ & 77.0 & 75.9 & 96.0 & 69.2 & 79.5 \\
  JiGen \cite{carlucci2019domain} & $\times$ & 79.4 & 75.3 & 96.0 & 71.4 & 80.5 \\
  EISNet \cite{wang2020learning}  & $\times$ & 81.9 & 76.4 & 95.9 & 74.3 & 82.2 \\
  MASF \cite{dou2019domain}       & $\times$ & 80.3 & 77.2 & 95.0 & 71.7 & 81.0 \\
  DAEL \cite{zhou2021domain}      & $\times$ & \underline{84.6} & 74.4 & 95.6 & 78.9 & 83.4 \\
  L2A-OT \cite{zhou2020learning}  & $\times$ & 83.3 & 78.2 & \underline{96.2} & 73.6 & 82.8 \\
  DDAIG \cite{zhou2020deep}       & $\times$ & 84.2 & 78.1 & 95.3 & 74.7 & 83.1 \\
  FACT \cite{xu2021fourier}       & $\times$ & \textbf{85.4} & 78.4 & 95.2 & 79.2 & 84.5 \\
  MixStyle \cite{zhou2021mixstyle} & $\times$ & 84.1 & 78.8 & 96.1 & 75.9 & 83.7 \\
  EFDMix \cite{zhang2022exact} & $\times$ & 83.9 & 79.4 & \textbf{96.8} & 75.0 & 83.9 \\
  DSU \cite{li2021uncertainty} & $\times$ & 83.6 & 79.6 & 95.8 & 77.6 & 84.1 \\
  StyleNeo \cite{kang2022style}   & $\times$ & 84.4 & 79.2 & 94.9 & 83.2 & 85.4 \\
  RSC \cite{huang2020self}        & $\times$ & 83.4 & 80.3 & 96.0 & 80.9 & 85.2 \\
  I$^{2}$-ADR \cite{meng2022attention}        & $\times$ & 82.9 & \underline{80.8} & 95.0 & 83.5 & 85.6 \\
  CDG \cite{du2022cross}        & $\times$ & 83.5 & 80.1 & 95.6 & 83.8 & \underline{85.8} \\
  \hline
  FedAvg \cite{pmlr-v54-mcmahan17a} & $\checkmark$ & 79.7 & 75.6 & 94.7 & 81.1 & 82.8 \\
  CASC \cite{yuan2023collaborative} & $\checkmark$ & 82.0 & 76.4 & 95.2 & 81.6 & 83.8 \\
  GA \cite{zhang2023federated} & $\checkmark$ & 83.2 & 76.9 & 94.0 & 82.9 & 84.3 \\
  ELCFS \cite{liu2021feddg} & $\checkmark$ & 82.3 & 74.7 & 93.3 & 82.7 & 83.2 \\
  CCST \cite{chen2023federated} & $\checkmark$ & 81.3 & 73.3 & 95.2 & 80.3 & 82.5 \\
  FADH \cite{xu2023federated} & $\checkmark$ & 83.8 & 77.2 & 94.4 & \underline{84.4} & 85.0 \\
  StableFDG \cite{park2023stablefdg} & $\checkmark$ & 83.0 & 79.3 & 94.9 & 79.8 & 84.2 \\
  COPA \cite{wu2021collaborative} & $\checkmark$ & 83.3 & 79.8 & 94.6 & 82.5 & 85.1 \\
  MCSAD (ours) & $\checkmark$ & 84.2 & \textbf{81.2} & 95.1 & \textbf{84.6} & \textbf{86.3} \\
  \hline
  \end{tabular}
  }
  }
  \caption{Accuracy(\%) on PACS dataset. We have bolded the best results and underlined the second results.}
\label{pacs}
\end{table}

\subsection{Comparison With the State-of-the-Art Methods}

\begin{table}[htbp]
\centering
  \scalebox{0.78}{
  \setlength{\tabcolsep}{1.0mm}{
  \begin{tabular}{lcccccc}
  \hline
  Methods & Dec. & Art & Clipart & Product &  Real  & Avg \\
  \hline
  DeepAll \cite{zhou2021domain}   & $\times$ & 57.9 & 52.7 & 73.5 & 74.8 & 64.7 \\
  JiGen \cite{carlucci2019domain} & $\times$ & 53.0 & 47.5 & 71.5 & 72.8 & 61.2 \\
  EISNet \cite{wang2020learning}  & $\times$ & 56.8 & 53.3 & 72.3 & 73.5 & 64.0 \\
  DAEL \cite{zhou2021domain}      & $\times$ & 59.4 & 55.1 & 74.0 & 75.7 & \underline{66.1} \\
  L2A-OT \cite{zhou2020learning}  & $\times$ & \textbf{60.6} & 50.1 & \underline{74.8} & \textbf{77.0} & 65.6 \\
  DDAIG \cite{zhou2020deep}       & $\times$ & 59.2 & 52.3 & 74.6 & \underline{76.0} & 65.5 \\
  FACT \cite{xu2021fourier}       & $\times$ & 60.3 & 54.9 & 74.5 & 76.6 & 66.6 \\
  MixStyle \cite{zhou2021mixstyle}& $\times$ & 58.7 & 53.4 & 74.2 & 75.9 & 65.5 \\
  StyleNeo \cite{kang2022style}   & $\times$ & 59.6 & 55.0 & 73.6 & 75.5 & 65.9 \\
  RSC \cite{huang2020self}        & $\times$ & 58.4 & 47.9 & 71.6 & 74.5 & 63.1 \\
  CDG \cite{du2022cross}          & $\times$ & 59.2 & 54.3 & 74.9 & 75.7 & 66.0 \\
  DomainDrop \cite{guo2023domaindrop} & $\times$ & 59.6 & 55.6 & 74.5 & 76.6 & 66.6 \\
  \hline
  FedAvg \cite{pmlr-v54-mcmahan17a}  & $\checkmark$  & 58.2 & 51.6 & 73.1 & 73.8 & 64.2 \\
  GA \cite{zhang2023federated}       & $\checkmark$ & 58.8 & 54.3 & 73.7 & 74.7 & 65.4 \\
  ELCFS \cite{liu2021feddg}          & $\checkmark$ & 57.8 & 54.9 & 71.1 & 73.1 & 64.2 \\
  CCST \cite{chen2023federated}      & $\checkmark$ & 59.1 & 50.1 & 73.0 & 71.7 & 63.6 \\
  FADH \cite{xu2023federated}        & $\checkmark$ & \underline{59.9} & 55.8 & 73.5 & 74.9 & 66.0 \\
  StableFDG \cite{park2023stablefdg} & $\checkmark$ & 57.2 & \underline{57.9} & 72.8 & 72.2 & 65.0 \\
  COPA \cite{wu2021collaborative}    & $\checkmark$ & 59.4 & 55.1 & \underline{74.8} & 75.0 & \underline{66.1} \\
  MCSAD (ours)                      & $\checkmark$ & 59.4 & \textbf{58.8} & \textbf{75.0} & 75.4 & \textbf{67.2} \\
  \hline
  \end{tabular}
  }
  }
  \caption{Accuracy(\%) on Office-Home dataset. We have bolded the best results and underlined the second results.}
\label{office_home}
\end{table}

\begin{table}[htbp]
\centering
  \scalebox{0.76}{
  \setlength{\tabcolsep}{1.0mm}{
  \begin{tabular}{lcccccc}
  \hline
  Methods & Dec. & Pascal & LabelMe & Caltech & Sun  & Avg \\
  \hline
  DeepAll \cite{zhou2021domain}   & $\times$ & 71.4 & 59.8 & 97.5 & 69.0 & 74.4 \\
  JiGen \cite{carlucci2019domain} & $\times$ & \underline{74.0} & 61.9 & 97.4 & 66.9 & 75.1 \\
  L2A-OT \cite{zhou2020learning}  & $\times$ & 72.8 & 59.8 & \textbf{98.0} & 70.9 & 75.4 \\
  MixStyle \cite{zhou2021mixstyle} & $\times$ & 72.6 & 58.5 & 97.7 & 73.3 & 75.5 \\
  RSC \cite{huang2020self}        & $\times$ & \textbf{75.3} & 59.8 & 97.0 & 71.5 & 75.9 \\
  DomainDrop \cite{guo2023domaindrop} & $\times$ & 76.4 & 64.0 & 98.9 & 73.7 & 78.3 \\
  \hline
  FedAvg \cite{pmlr-v54-mcmahan17a} & $\checkmark$ & 72.0 & 63.3 & 96.5 & 72.4 & 76.0 \\
  CASC \cite{yuan2023collaborative} & $\checkmark$ & 72.0 & \underline{63.5} & 97.2 & 72.1 & \underline{76.2} \\
  ELCFS \cite{liu2021feddg} & $\checkmark$ & 71.1 & 59.5 & 96.6 & \underline{74.0} & 75.3 \\
  COPA \cite{wu2021collaborative} & $\checkmark$ & 71.5 & 61.0 & 93.8 & 71.7 & 74.5 \\
  MCSAD (ours) & $\checkmark$ & 76.1 & \textbf{65.6} & \underline{98.6} & \textbf{75.0} & \textbf{78.8} \\
  \hline
  \end{tabular}
  }
  }
  \caption{Accuracy(\%) on VLCS dataset. We have bolded the best results and underlined the second results.}
\label{vlcs}
\end{table}

As shown in Table \ref{pacs}, Table \ref{office_home}, and Table \ref{vlcs}, our method MCSAD achieves the best average accuracy under the data decentralization scenario on three datasets, PACS, Office-Home, and VLCS. Compared with the style interpolation methods, e.g. ELCFS and CCST, our method can explore the broader style space, which leads to large performance improvement. FADH trains the additional image generators on each source domain by maximizing the entropy of the global model and minimizing the cross-entropy of the local model. Compared with FADH, our method MCSAD is simple but effective. As shown in Table \ref{pacs} and Table \ref{office_home}, our method outperforms FADH significantly. COPA learns the domain-invariant model with the collaboration of multiple classifier heads, and the ensemble of classifier heads from different domains is used to deploy on the unseen domain. However, our method only utilizes a global model to deploy on unseen domain and achieves better performance, as shown in Table \ref{pacs}, Table \ref{office_home}, and Table \ref{vlcs}. Compared with the methods which focus on the model aggregation stage, e.g. CASC and GA, our method learns the domain-invariant representations on local clients and outperforms these methods largely.

Although the data from different domains are kept decentralized, our method even outperforms the state-of-the-art methods e.g. MixStyle, RSC, and StyleNeo, which can access multiple source domain data simultaneously.

\subsection{Ablation Study}

\textbf{Contributions of Different Components.} As shown in Table \ref{contri_pacs}, we conduct ablation study about the different components of our method on PACS dataste. Compared with the baseline, the Collaborative Style Augmentation (CSA) can improve the average accuracy largely. Combining with the $\mathcal{L}_{SupCon}$ and $\mathcal{L}_{cdrm}$, the performance can be further improved by learning the compact representations and the class relationship. By alternative conduct style augmentation and semantic learning with $\mathcal{L}_{SupCon}$ and $\mathcal{L}_{cdrm}$, we achieve the best average accuracy.

The CSA proposed in this work augments the feature in hidden layer. We also conduct experiments to validate the effectiveness of conducting CSA on different layers. As shown in Table \ref{blocks_pacs}, we conduct CSA after the different blocks of ResNet-18. Compared with the baseline method, conducting CSA after different blocks can improve the accuracy on unseen target domain. Due to the shallow layer of Deep Neural Network extracting the features with more style information, conducting CSA after the first and second Blocks can achieve better accuracy. When conducting CSA after the first and second Blocks, we can achieve the best average accuracy 85.0\%.

\textbf{Comparison with Different Style Augmentation Methods.} We also make a comparison with other data augmentation methods under the FedAvg framework, including (1) single-domain style expanding methods, e.g. DSU and AdvStyle, (2) multi-domain style interpolation method, e.g. AM, MixStyle, and EFDMix. (3) pre-defined augmentation pool, e.g. RandAug, and (4) the methods FADH, which train image generator.

\begin{table}[htbp]
\centering
\scalebox{0.78}{
\setlength{\tabcolsep}{2.0mm}{
\begin{tabular}{cccccccc}
\hline
CSA & $\mathcal{L}_{SupCon}$ & $\mathcal{L}_{cdrm}$ & Art & Cartoon & Photo & Sketch & Avg \\
\hline
 - & - & - & 79.7 & 75.6 & 94.7 & 81.1 & 82.8 \\
$\checkmark$ & - & - & 82.3 & 80.2 & 95.4 & 82.2 & 85.0 \\
$\checkmark$ & $\checkmark$ & - & 83.4 & 80.9 & \textbf{95.7} & 83.9 & 86.0 \\
$\checkmark$ & - & $\checkmark$ & 82.5 & 80.1 & 95.5 & 83.6 & 85.4 \\
$\checkmark$ & $\checkmark$ & $\checkmark$ & \textbf{84.2} & \textbf{81.2} & 95.1 & \textbf{84.6} & \textbf{86.3} \\
\hline
\end{tabular}
}
}
\caption{Accuracy(\%) of each component on PACS dataset.}
\label{contri_pacs}
\end{table}

\begin{table}[htbp]
\centering
\scalebox{0.78}{
\setlength{\tabcolsep}{2.0mm}{
\begin{tabular}{cccccccc}
\hline
Block1 & Block2 & Block3 & Art & Cartoon & Photo & Sketch & Avg \\
\hline
- & - & - & 79.7 & 75.6 & 94.7 & 81.1 & 82.8 \\
$\checkmark$ & - & - & 81.4 & 78.3 & 95.1 & 83.0 & 84.5 \\
- & $\checkmark$ & - & 81.7 & 79.8 & \textbf{95.4} & 81.3 & 84.6 \\
- & - & $\checkmark$ & \textbf{82.7} & 79.3 & 95.3 & 80.3 & 84.4 \\
$\checkmark$ & $\checkmark$ & - & 82.3 & \textbf{80.2} & \textbf{95.4} & 82.2 & \textbf{85.0} \\
$\checkmark$ & $\checkmark$ & $\checkmark$ & 80.9 & 78.1 & 94.7 & \textbf{83.2} & 84.2 \\
\hline
\end{tabular}
}
}
\caption{Ablation study about the location of collaborative style augmentation on PACS dataset.}
\label{blocks_pacs}
\end{table}

\begin{table}[htbp]
\centering
\scalebox{0.78}{
  \setlength{\tabcolsep}{1.5mm}{
  \begin{tabular}{lcccccc}
  \hline
  Methods & Art & Cartoon & Photo & Sketch & Avg \\
  \hline
  FedAvg \cite{pmlr-v54-mcmahan17a}    & 79.7 & 75.6 & 94.7 & 81.1 & 82.8 \\
  + DSU \cite{li2021uncertainty}            & 81.5 & 77.9 & 95.5 & 81.6 & 84.1 \\
  + AdvStyle \cite{zhong2022adversarial} & 80.0	& 75.9 & 94.7 & 82.0 & 83.2 \\
  + AM \cite{zhang2023federated}       & 83.4 & 76.2 & 95.4 & 80.8 & 84.0 \\
  + MixStyle \cite{zhou2021mixstyle}   & 82.1 & 76.8 & 95.3 & 82.3 & 84.1 \\
  + EFDMix \cite{zhang2022exact}       & 81.9 & 78.3 & 94.7 & 82.5 & 84.4 \\
  + RanAug \cite{wu2021collaborative}  & \textbf{83.6} & 76.1 & \textbf{95.9} & 81.4 & 84.3 \\
  + FADH \cite{xu2023federated}        & 82.8 & 77.2 & 93.9 & \textbf{84.1} & 84.5 \\
  + CSA (ours)                      & 82.3 & \textbf{80.2} & 95.4 & 82.2 & \textbf{85.0} \\
  \hline
  \end{tabular}
  }
}
\caption{Accuracy (\%) of different style augmentation methods on PACS dataset.}
\label{augments_pacs}
\end{table}

As shown in Table \ref{augments_pacs}, the CSA method proposed in this work outperforms the single-domain style expanding methods and multi-domain style interpolation methods on average accuracy largely. Our method can generate novel styles out of the existing source domains with the collaboration of multiple classifier heads, the single-domain style expanding methods or multi-domain style interpolation methods only lead to limited styles in the style space of the existing source domains.

The RandAug generates images with pre-defined augmentation strategies, e.g. image processing, which cannot dynamically generate novel styles out of the existing source domains. As shown in Table \ref{augments_pacs}, RandAug performs poorly on the Sketch domain. Different from RandAug, our CSA method generates novel styles by adversarial style augmentation, which can explore the out-of-distribution styles in an online manner. As shown in Table \ref{augments_pacs}, our CSA also outperforms RandAug in average accuracy.

The FADH trains the image generators on multiple decentralized source domains and achieves the average accuracy 84.5\%. Different from FADH, our method CSA diversify the features by adversarial style augmentation, which leads to better average accuracy. Furthermore, compared with FADH, our method leads to smaller computational and storage cost.

\textbf{Different Distillation Strategies.} In the domain-invariant learning stage, we propose Cross-Domain Relation Matching (CDRM) $\mathcal{L}_{cdrm}$ loss to distill the intrinsic relation between classes.

There are different distillation strategies, including (1) Kullback-Leibler (KL) divergence used by \cite{kang2022style,xu2021fourier}, hear we align the logits of augmented features from current model to the ensembled logits of original features from multiple classifier heads, (2) Collaboration of Frozen Classifiers (CoFC) used by COPA \cite{wu2021collaborative}, the CoFC freezes the classifier heads from other domains to learn the domain-invariant feature extractor. (3) Cross-Domain Relation Matching (CDRM) proposed in this work, the class relation from original feature and augmented feature are aligned to the ensembled classes relation, as shown in Equation \ref{eq11}.

\begin{table}[htbp]
\centering
\scalebox{0.78}{
\setlength{\tabcolsep}{1.6mm}{
\begin{tabular}{l c c c c c}
\hline
Method & Art & Cartoon & Photo & Sketch & Avg \\
\hline
Baseline & 82.3 & \textbf{80.2} & 95.4 & 82.2 & 85.0 \\
+ KL \cite{kang2022style} & 82.1 & 78.9 & 94.5 & \textbf{84.5} & 85.0 \\
+ CoFC \cite{wu2021collaborative} & 81.2 & 78.2 & 95.1 & 84.1 & 84.7 \\
+ CDRM (ours) & \textbf{82.5} & 80.1 & \textbf{95.5} & 83.6 & \textbf{85.4} \\
\hline
\end{tabular}
}
}
\caption{Accuracy (\%) of different distillation strategies.}
\label{align_pacs}
\end{table}

\begin{figure}[htb]
    \centering
    \includegraphics[width=0.48\textwidth]{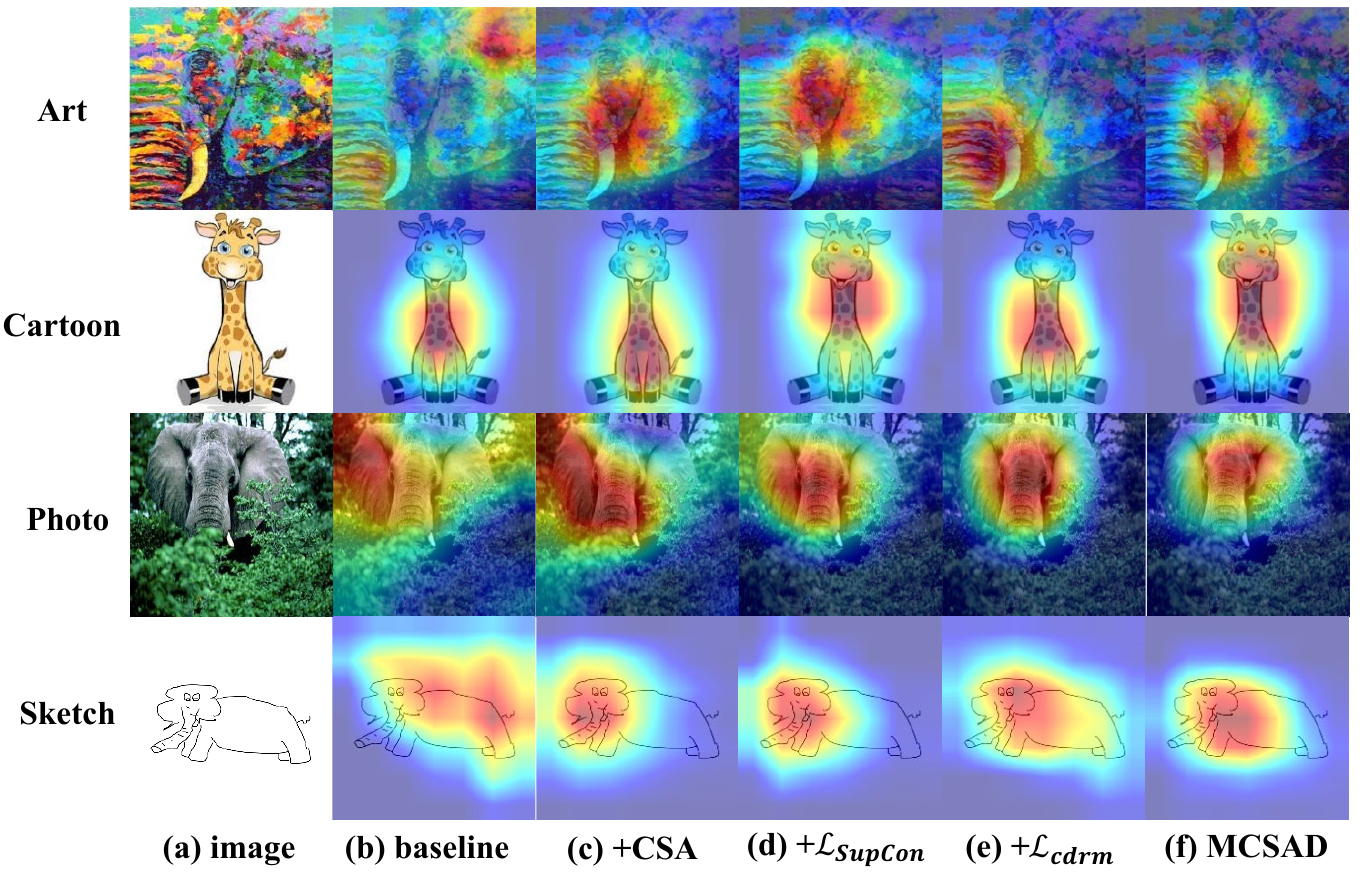}
    \caption{Visualization by Grad-CAM on unseen target domain.}
    \label{cam}
\end{figure}

As shown in Table \ref{align_pacs}, we report the results of different distillation strategies by combining the CSA method proposed in this work. The KL lead to limited improvement on the average accuracy, and the CoFC even leads to degraded performance on average accuracy. On the Art and Photo domains, KL and CoFC cannot achieve improvement. The CDRM strategy proposed in this work outperforms KL and CoFC significantly by distilling the relationship between classes to improve the generalization ability of model.

\textbf{Visualization.} Furthermore, we also visualize the activated region on input images for classification by Grad-CAM \cite{selvaraju2017grad}. As shown in Figure \ref{cam}, the learned baseline model usually focuses on the texture or background region on the unseen domain. By using CSA, the learned model can capture the region of the object for classification. Combined with $\mathcal{L}_{SupCon}$ and $\mathcal{L}_{cdrm}$, the learned model tends to capture the discriminative and overall region of the object, which leads to better performance on the unseen domain. By combining all components, the most discriminative region can be focused on for better generalization performance on unseen domain.

\section{Conclusion}

In this paper, we propose a MCSAD method to solve the multi-source domain generalization problem under the data decentralization scenario. To explore the out-of-distribution styles on the decentralized source domains, we propose a multi-source collaborative style augmentation method to generate features with novel styles for diversifying the source domains. Moreover, we propose the domain-invariant learning between the original data and augmented data to learn the domain-invariant representations and improve the generalization ability of model. The extensive experiments can validate the effectiveness of the proposed MCSAD method.

\bibliographystyle{named}
\bibliography{ijcai25}

\end{document}